\documentclass[11pt]{article}
\usepackage[preprint]{acl}
\usepackage[T1]{fontenc}
\usepackage{lmodern}
\usepackage{latexsym}
\usepackage{amsmath,amssymb}
\usepackage{booktabs}
\usepackage{array}
\usepackage{graphicx}
\usepackage{microtype}

\title{Data Quality over Capacity: Internalizing Documents into LoRA
Adapters for Closed-Book QA}
\author{Joan Figuerola Hurtado \\ Independent Researcher}
\date{}
\begin{document}
\maketitle
\begin{abstract}
We study baking documents directly into the weights of a 4-bit
Gemma-4-e4b model via LoRA, so a system can answer questions about a
corpus \emph{closed-book}: no retrieval and no context-window budget.
Across roughly 100 training runs from single documents to a 99-document
corpus, we find that once adapter capacity is adequate,
\textbf{training-data quality is the dominant lever} on closed-book
accuracy, outweighing LoRA rank, learning rate, and two alternative
architectures combined; capacity itself is a hard gate below which no
data intervention helps. A single curation pass (shortening gold answers
to canonical 1--6 word spans and dropping trivia) moved closed-book
accuracy from 57.7\% to 85.7\% on a 15-document corpus, a larger jump
than any architectural change. We confirm a capacity trend (rank must
grow with corpus size) entangled with a coupling between rank and
learning rate that we initially misdiagnosed. On a 15-document slice we
add a real retrieval baseline: the internalized adapter (84.2\% recall)
beats a BM25-RAG pipeline with a base reader (58.9\%) and even a
realistic gold-chunk oracle (65.6\%) at lower latency. We report the
full arc, including three misdiagnoses, as a case study in debugging LLM
training empirically.
\end{abstract}
\section{Introduction}\label{introduction}

Retrieval-augmented generation (RAG) is the default way a language model
uses a corpus it was not trained on: retrieve relevant chunks and place
them in context. This caps how much of a corpus informs a single answer,
adds retrieval as a failure mode, and spends context budget on every
query. An alternative is \textbf{knowledge internalization}: train a
lightweight LoRA adapter \citep{hu2022lora} to answer questions about a
corpus \emph{closed-book}, with no document text in the prompt. Prior
work weighs fine-tuning against retrieval for injecting new knowledge
\citep{ovadia2024finetuning} and finds that supervised fine-tuning on
new facts can be slow and hallucination-prone
\citep{gekhman2024finetuning}, while a parallel line quantifies how much
factual knowledge a model's parameters can store
\citep{allenzhu2024storage}. Our north star is to internalize a document
collection directly into a model's weights, so its content can be
searched and queried without a retrieval step and without the fixed
context-window limit, even as the collection grows to tens of thousands
of documents.

We report roughly two months of experimentation: what recipe reliably
answers closed-book questions about documents an adapter ``read'' during
training, and how it scales from 1 to 99 documents. Our headline is
unglamorous but, we think, generalizable:
\textbf{once the adapter has enough capacity, the biggest lever is not further capacity or recipe, but the mundane quality of the training QA data},
specifically how verbose the gold answers are.

\textbf{Contributions.} (1) A worked example where a training-data
curation pass yields a bigger gain (+28 points on 15 documents) than
architecture changes (EntiGraph-style synthetic pretraining
\citep{yang2024entigraph}; trainable KV-cache cartridges
\citep{eyuboglu2025cartridges}). (2) A corpus-scaling ladder isolating a
rank-must-grow-with-corpus trend, and a transferable lesson
(\emph{larger rank needs a gentler learning rate}) that first
masqueraded as a capacity ceiling. (3) A real BM25-RAG baseline showing
internalization beats retrieval on this task and hardware, with an
honest metric caveat. All numbers are single-seed; we treat that and a
style-biased metric as the central threats (\S\ref{sec:discussion}).

\section{Task, Metric, and Recipe}\label{task-metric-and-recipe}

\label{sec:setup}

\textbf{Task.} For a document corpus we generate closed-book
question--answer pairs whose gold answers are grounded verbatim in the
source. We train an adapter to answer \emph{without} the source in the
prompt. Training data is self-generated by a strong LLM reading each
document's own extracted chunks (no dedicated QA-teacher model); every
claim carries a verbatim source quote and 8 training
paraphrase-questions plus 2 held-out questions.

\textbf{Metric.} A prediction is correct on exact substring match
(either direction) or token-F1 \ensuremath{\geq} 0.74. We report
\textbf{recall} (known-answer questions) separately from
\textbf{overall} accuracy (which also scores an abstention slice for
questions about never-trained facts). The metric is lenient in absolute
terms and structurally favors short gold answers; we test this directly
in \S\ref{sec:results} and Appendix \ref{app:metric}.

\textbf{Recipe.} Base model \texttt{gemma-4-e4b-it-4bit}, LoRA on MLP
projections only across all 42 layers, trained via MLX on Apple Silicon.
Two stages resumed into one adapter: \textbf{Stage A} (CPT-like)
completion-loss training on raw chunks plus multi-level paraphrases
\citep{maini2024wrap}, then \textbf{Stage B} (SFT) closed-book QA
supervision, answer-token loss only. Full hyperparameters and cost in
Appendix \ref{app:config}.

\section{Results}\label{results}

\label{sec:results}

\textbf{The ceiling, and the floor.} On a single document a modest LoRA
(rank 32, 3k steps) reaches 86--98\% closed-book overall accuracy; two
documents, 94\%. A base model with no context scores
\ensuremath{\sim}0\%. The interesting regime is many documents.

\textbf{Data quality is the dominant lever.} At 15 documents,
paid-teacher data reaches 78.9\% overall (76.3\% recall). Swapping to
self-generated claims regressed sharply to 57.7\% (55.6\% recall).
Ablations ruled out capacity (rank 32\ensuremath{\rightarrow}64: +3
pts), recipe (single-stage was \emph{worse}), factual quality, and
under-training. The cause was data \emph{learnability}: the
self-generated model could not fit its own data below validation loss
0.287 (vs.~0.046 for teacher data); the pipeline had over-mined
high-entropy trivia and phrased answers as long clauses. Re-curating
(short canonical answers, dropped trivia, de-duplication) dropped the
loss floor to 0.056 and raised accuracy to \textbf{85.7\%}, a
\textbf{+28-point} gain from data alone (Table \ref{tab:main}), larger
than any architectural or capacity intervention, and replicated at
99-document scale (below).

Because curation shortens answers and our metric favors short answers,
we checked whether this is a scoring artifact by re-scoring 3,792 saved
predictions under stricter rules. It is not: the
answer-length\ensuremath{\rightarrow}recall gradient \emph{widens} under
exact match (from +30 to +47 points), because lenient matching grants
\emph{more} partial credit to long answers (Appendix \ref{app:metric}).
The residual confound, longer answers being intrinsically harder to
produce, is left to a controlled ablation.

\textbf{Architectures do not beat curated LoRA.} On the same corpus,
EntiGraph-style synthetic-CPT augmentation \citep{yang2024entigraph}
moved overall accuracy 84.0\% \ensuremath{\rightarrow} 82.1\%, a null
within single-seed noise (\ensuremath{\sim}1.7 pts across matched runs).
A from-scratch Cartridges \citep{eyuboglu2025cartridges} (trainable
KV-cache prefix) implementation reached only 11.9\% (cross-entropy) and
3.78\% (with context-distillation), with the distillation variant's
abstention collapsing to 0\%; we report this as
\emph{our implementation did not become competitive}, not an
architectural verdict. Neither earns its complexity at this scale.

\begin{table}[t]
\centering\small
\begin{tabular}{@{}lrr@{}}
\toprule
15-doc setting (rank 64) & Recall & Overall \\
\midrule
teacher data (1-stage) & 76.3\% & 78.9\% \\
self-generated (2-stage) & 55.6\% & 57.7\% \\
\textbf{+ curation} & \textbf{84.2\%} & \textbf{85.7\%} \\
\;\; + EntiGraph augmentation & 82.2\% & 82.1\% \\
\bottomrule
\end{tabular}
\caption{Data curation is the largest single lever (+28 points), exceeding any architecture change. All single-seed; N=1,772--3,990 held-out questions.}
\label{tab:main}
\end{table}

\textbf{Capacity must scale with corpus size.} On a nested corpus (CL-15
\ensuremath{\subset} CL-25 \ensuremath{\subset} CL-50
\ensuremath{\subset} CL-100), rank-64 recall collapses from 71\% at 25
docs to 50\% at 50 docs, with flat cross-doc mis-binding and a rising
loss floor, the signature of \emph{capacity saturation}
\citep{allenzhu2024capacity,back2026lora} rather than interference.
Doubling to rank 128 restores it (Table \ref{tab:ladder}). We frame this
as a trend, not a law: it rests on two rank-doublings and, importantly,
the ladder held learning rate fixed, so rank and LR-adequacy are only
partly separated.

\textbf{Larger rank needs a gentler learning rate.} Doubling again to
rank 256 at CL-100, reusing the rank-128 learning rate, \emph{regressed}
recall (46.7\% vs.~55.2\%). We first blamed a sequence-length confound;
directly counting truncated tokens falsified that in minutes (zero QA
rows truncated). The real cause was an un-retuned learning rate: halving
it (4e-5\ensuremath{\rightarrow}2e-5) dropped the loss floor and
recovered recall to 61.6\%, beating rank-128 by 6 points. Two
plausible-but-wrong diagnoses were distinguished from the real one only
by direct measurement.

\begin{table}[t]
\centering\small
\begin{tabular}{@{}llrr@{}}
\toprule
Tier & Rank (LR) & Overall & Recall \\
\midrule
CL-25 & 64 (4e-5) & 74.6\% & 71.1\% \\
CL-50 & 64 (4e-5) & 56.5\% & 50.0\% \\
CL-50 & 128 (4e-5) & 68.7\% & 64.0\% \\
CL-100 & 128 (4e-5) & 60.7\% & 55.2\% \\
CL-100 & 256 (4e-5) & 53.3\% & 46.7\% \\
CL-100 & 256 (\textbf{2e-5}) & 66.3\% & \textbf{61.6\%} \\
\bottomrule
\end{tabular}
\caption{Rank must grow with corpus size (r64 saturates 25--50 docs); and larger rank needs a gentler LR (the r256 regression is an LR artifact, not capacity). Single-seed; corpus recall.}
\label{tab:ladder}
\end{table}

\textbf{Closing the loop.} At CL-100, an older document cohort lagged a
newer one; the only measured difference was gold-answer length (7.5
vs.~4.1 words), a consequence of the answer-brevity rule being specified
partway through construction. Re-authoring the older 51 documents under
the current spec raised overall recall from 61.6\% to \textbf{68.4\%},
converging all cohorts, and replicating the data-quality finding at
\ensuremath{\sim}7\ensuremath{\times} the original corpus scale.

\textbf{A real retrieval baseline.} Our motivation is that
internalization replaces retrieval, yet a prior ``oracle-RAG'' number
(\ensuremath{\sim}100\%) turned out to force-inject the gold answer and
copy it, a trivial upper bound rather than retrieval. We built a genuine
baseline on the 15-document corpus, evaluated on the same 2,009
questions and scorer: a BM25 retriever over the raw chunks feeding a
\emph{base} reader (no adapter), plus a realistic oracle that feeds the
true gold chunk to that reader (Table \ref{tab:rag}). The internalized
adapter (84.2\% recall) beats BM25-RAG by \ensuremath{\sim}25 points and
even the realistic oracle by \ensuremath{\sim}19 points, at a fraction
of the latency. BM25 surfaces the gold chunk only 53\% of the time at
rank 1 (86\% at 5), and even with it present the base reader reaches
only 65.6\%: retrieval and reading are both lossy.

\begin{table*}[t]
\centering\small
\begin{tabular}{@{}lllrrr@{}}
\toprule
Arm & Reader & Context & Recall & Overall & Gen.\ latency p50 \\
\midrule
base, no context (floor) & base & none & \ensuremath{\sim}0\% & \ensuremath{\sim}0\% & --- \\
BM25-RAG, $k{=}5$ & base & top-5 retrieved chunks & 58.9\% & 62.6\% & 2,895 ms \\
realistic oracle & base & gold document chunk & 65.6\% & 67.5\% & 672 ms \\
\textbf{closed-book adapter} & +LoRA & internalized (none in prompt) & \textbf{84.2\%} & \textbf{85.7\%} & \ensuremath{\sim}300--800 ms \\
copy-task "oracle" (prior) & base & gold \emph{answer} injected & \ensuremath{\sim}100\% & 99.9\% & --- \\
\bottomrule
\end{tabular}
\caption{Real retrieval baseline on the 15-document corpus (N=2,009; same deterministic scorer). Internalization beats BM25-RAG and a realistic gold-chunk oracle on accuracy and latency. Caveat: the metric favors the style-matched adapter, so RAG's accuracy is a lower bound on its knowledge (\S\ref{sec:discussion}).}
\label{tab:rag}
\end{table*}

\section{Discussion and Threats to
Validity}\label{discussion-and-threats-to-validity}

\label{sec:discussion}

\textbf{Unified findings.} (1) Capacity gates; once adequate, data
quality dominates, and the two are complementary rather than competing.
(2) Rank must grow with corpus size
(\ensuremath{\sim}2\ensuremath{\times} rank per
\ensuremath{\sim}2\ensuremath{\times} docs, over two doublings). (3)
Larger rank needs a gentler learning rate; failing to retune mimics a
capacity regression. (4) Interference (rising mis-binding, stable loss
floor) and capacity saturation (uniform collapse, rising floor) have
distinguishable signatures, which picks the right fix. (5) Two
architectural alternatives did not beat curated LoRA at this scale in
our hands.

\textbf{Threats.} \emph{Single seed}: every run uses seed 42;
sub-2-point differences are within observed noise, and load-bearing
deltas need multi-seed replication we have not done.
\emph{Metric favors the adapter}: the deterministic scorer rewards the
style-matched adapter over a paraphrasing base RAG reader, so RAG's
accuracy is a lower bound on its knowledge; the reader-independent BM25
hit-rate (86\% at \(k{=}5\)) partly controls for this, and an LLM-judge
re-scoring is the fair follow-up. \emph{Scale}: 99 documents is far
below a tens-of-thousands-document target, so the negative architecture
results and the exact rank/LR constants may not extrapolate;
larger-scale cartridge training in particular reports gains our
small-scale runs cannot
\citep{eyuboglu2025cartridges,hardalov2026cartridgesatscale}.
\emph{Single model/hardware and a 4-bit base}: quantization could set
part of the loss floor we attribute to data; a higher-precision control
is needed.

\section{Conclusion}\label{conclusion}

The strongest and most reproducible finding across \ensuremath{\sim}100
runs is that training-data quality, specifically answer brevity and
removing trivia, is a larger and more reliable lever for closed-book
internalization than capacity, learning-rate tuning, or the alternative
architectures we tested, provided capacity is adequate. Internalization
also beats a real BM25-RAG baseline and a realistic oracle on accuracy
and latency at this scale, modulo a metric that favors the fine-tuned
reader. Future work: multi-seed variance, an LLM-judge RAG comparison at
full corpus scale, and routing across multiple smaller adapters toward a
tens-of-thousands-document target.

\appendix

\section{Full Configuration and Cost}\label{full-configuration-and-cost}

\label{app:config}

Two stages resumed into one adapter (values common unless noted). Base
\texttt{mlx-community/gemma-4-e4b-it-4bit}; LoRA MLP-only
(\texttt{up/down/gate\_proj}), 42 layers, alpha 20.0, dropout 0.05;
AdamW; effective batch 8 (batch 1 \ensuremath{\times} grad-accum 8);
cosine LR, warmup-init 1e-7, decay to 0.1\ensuremath{\times} peak; peak
LR 4e-5 (2e-5 for rank 256); Stage-B steps 12k/21k/40.5k/85.2k for
CL-15/25/50/100; max sequence length 3,072 (2,048 for rank 256);
answer-token loss masking; seed 42.

Trainable parameters and adapter size: rank 64 = 103M (413 MB), rank 128
= 206M (826 MB), rank 256 = 413M (1.65 GB); cartridge prefix 29 MB.
Stage-B wall-clock \ensuremath{\sim}1--1.5 h (15 docs) to several hours
(CL-100). Inference latency was not separately profiled for the adapter
(the budget targets p50 450 ms).

\section{Metric-Bias Analysis}\label{metric-bias-analysis}

\label{app:metric}

Re-scoring 3,792 saved known-answer predictions (15-doc runs) reproduces
the reported recall exactly, confirming the re-scoring is faithful.
Lenient matching inflates absolute pass rates by \ensuremath{\sim}11--14
points over exact match (e.g.~self-generated: 55.6\% shipped vs.~41.8\%
exact). But the answer-length\ensuremath{\rightarrow}recall gradient
\emph{widens} under exact match (1-word vs.~13+-word gold: +30 points
shipped, +47 exact), because lenient rules grant more partial credit to
\emph{long} answers (a short prediction being a fragment of a long gold
answer). So the brevity effect is not a scoring artifact; if anything
the metric understates it. The residual confound is that longer answers
are intrinsically harder to reproduce, which a canonicalization-only
ablation on a fixed question set would isolate.

\section{Reproducibility Notes}\label{reproducibility-notes}

\label{app:repro}

The 85.7\% curated result and the 84.0/82.1 EntiGraph pair were re-run
from saved adapters and reproduced to within 0.1 point (85.66 / 83.97 /
82.08\% overall). The CL-100 rank-128 per-cohort recalls were re-derived
from committed per-row outputs under a
document\ensuremath{\rightarrow}cohort mapping validated by reproducing
the rank-256 and re-authored runs exactly: original-15 49.4\%, 16--25
52.8\%, 26--50 48.0\%, 51--99 60.2\% (corpus 55.2\%). All eval JSONs and
harness scripts are retained alongside the source.
\end{document}